\documentclass{article}

\usepackage[preprint]{neurips_2025}

\usepackage[utf8]{inputenc} 
\usepackage[T1]{fontenc}    
\usepackage{url}            
\usepackage{booktabs}       
\usepackage{amsfonts}       
\usepackage{nicefrac}       
\usepackage{microtype}      
\usepackage{xcolor}         

\usepackage{multirow}
\usepackage{float}
\usepackage[utf8]{inputenc} 
\usepackage[T1]{fontenc}    
\usepackage{url}            
\usepackage{booktabs}       
\usepackage{amsfonts}       
\usepackage{nicefrac}       
\usepackage{microtype}      
\usepackage{mathtools}
\usepackage{natbib}
\usepackage{xcolor}
\usepackage{url}
\usepackage{amsmath}
\usepackage{amssymb}
\usepackage{amsthm}
\usepackage{graphicx}
\usepackage{nicefrac}
\usepackage{appendix}
\usepackage{enumitem}
\usepackage{xspace}
\usepackage{pifont}
\usepackage{tikz}
\usetikzlibrary{positioning, arrows.meta, shapes, calc}

\usepackage{wrapfig}
\usepackage{caption}
\usepackage{subcaption}
\usepackage{array}
\definecolor{darker}{rgb}{0,0.15,0.65}
\usepackage[colorlinks, urlcolor=darker, citecolor=darker, linkcolor=darker]{hyperref} 

\usepackage{colortbl}
\definecolor{green}{HTML}{009B55}
\newcommand{\ours}{{DMA}\xspace}

\title{\ours: Online RAG Alignment with Human Feedback}

\author{%
  Yu Bai$^{1}$ \quad
  Yukai Miao$^{1}$ \quad
  Dawei Wang$^{1}$ \quad
  Li Chen$^{1}$ \quad
  Fei Long$^{2}$ \quad
  Rundi Zhai$^{4}$ \\
  \textbf{Dan Li}$^{2}$ \quad
  \textbf{Yanyu Ren}$^{2}$ \quad
  \textbf{Tianfeng Liu}$^{1}$ \quad
  \textbf{Hongtao Xie}$^{3}$ \quad
  \textbf{Ce Yang}$^{3}$ \quad
  \textbf{Xuhui Cai}$^{3}$ \\
  \\
  $^{1}$Zhongguancun Laboratory \\
  $^{2}$Tsinghua University \\
  $^{3}$China Mobile Communications Group Co., Ltd. \\
  $^{4}$Beijing University of Posts and Telecommunications \\ 
}

\begin{document}

\maketitle

\begin{abstract}
Retrieval-augmented generation (RAG) systems often rely on static retrieval, limiting adaptation to evolving intent and content drift. We introduce \emph{Dynamic Memory Alignment} (\ours), an online learning framework that systematically incorporates multi-granularity human feedback to align ranking in interactive settings. \ours organizes \emph{document-, list-, and response-level} signals into a coherent learning pipeline: supervised training for pointwise and listwise rankers, policy optimization driven by response-level preferences, and knowledge distillation into a lightweight scorer for low-latency serving. Throughout this paper, \emph{memory} refers to the model’s \emph{working memory}, which is the entire context visible to the LLM for In-Context Learning.

We adopt a dual-track evaluation protocol mirroring deployment: (i) large-scale online A/B ablations to isolate the utility of each feedback source, and (ii) few-shot offline tests on knowledge-intensive benchmarks. Online, a multi-month industrial deployment further shows substantial improvements in human engagement. Offline, \ours preserves competitive foundational retrieval while yielding notable gains on conversational QA (TriviaQA, HotpotQA). Taken together, these results position \ours as a principled approach to feedback-driven, real-time adaptation in RAG without sacrificing baseline capability.

\end{abstract}

\section{Introduction}
\label{sec:introduction}

RAG has become a central paradigm for improving the factuality and adaptability of large language models (LLMs) in knowledge-intensive tasks \citep{lewis2020retrieval,borgeaud2022improving}. By decoupling parametric memory from non-parametric retrieval, RAG enables models to access external information on demand, grounding responses in up-to-date, domain-specific knowledge without changing model parameters. This separation has underpinned recent progress in open-domain QA \citep{izacard2021leveraging}, multi-hop reasoning \citep{yang2018hotpotqa}, and instruction-based augmentation \citep{lin2024radit,gao2023retrieval}.

Despite these advances, deployed RAG systems face persistent challenges in dynamic online settings:  
(i) \textbf{Static retrieval} fails to adapt to evolving human intent and content drift; most dense retrievers are trained offline and remain fixed at deployment, underutilizing live interaction signals \citep{lin2023train,jiang2024mixtral}.  
(ii) \textbf{Context bottlenecks} in mainstream LLMs \citep{liu2023lost} force aggressive selection; relying only on top-$k$ embedding similarity yields suboptimal recall and necessitates principled reranking \citep{monot5,glass2022re2g,qin2023large}.  
(iii) \textbf{Limited online flexibility} of rankers/hybrid retrievers reduces personalization and real-time adaptation \citep{repllama,contriever,zhang2024arl2}.  
Together, these issues point to the need for an adaptive interface between \emph{human feedback} and \emph{retrieval control}.

We address this gap by framing it within the emerging discipline of Context Engineering—the task of optimally curating the information provided to an LLM. We posit that \emph{memory} is not the static corpus, but the LLM’s dynamic \emph{working memory}: the complete, aggregated context assembled at each turn. This working memory is a finite, high-leverage resource that directly determines what the model can condition on.

While current Context Engineering relies heavily on static heuristics or manual prompt design, the core challenge of dynamically aligning this context to evolving user intent remains unsolved. \ours provides a formal solution, offering a framework to learn the control of this resource. It moves beyond static curation to align the \emph{selection and ordering} of context with human preferences in real time.

\ours targets online adaptability through three components:  
(1) a \textbf{feedback taxonomy} tailored to interactive RAG, covering \emph{document-, list-, and response-level} signals;  
(2) \textbf{preference/reward modeling} that converts heterogeneous signals into structured supervision for pointwise and listwise policies;  
(3) \textbf{online fusion \& distillation} that aggregates teachers and serves a lightweight scorer under tight latency budgets.

This design makes \ours particularly suitable for human-facing applications—chat assistants, enterprise QA, and developer support—where responsiveness and continual alignment are critical \citep{asai2024reliable,jeong2024adaptive}.

\paragraph{Contributions.}  
\begin{itemize}[leftmargin=*, itemsep=0.3em, topsep=0em]
  \item \textbf{Framework.} We introduce \ours, an online learning framework that continuously refines retrieval and reranking from \emph{multi-level} human feedback. The approach unifies document-, list-, and response-level signals into a single alignment pipeline.
  \item \textbf{Evaluation protocol.} We adopt a dual-track evaluation: large-scale online A/B ablations to isolate each feedback source’s contribution, and few-shot offline tests on public knowledge-intensive benchmarks. \ours attains strong results on conversational QA (TriviaQA, HotpotQA), indicating suitability for dialogue-centric use cases.
  \item \textbf{Real-world impact.} In a multi-month randomized controlled deployment, \ours improves session-level human satisfaction by \textbf{+15.26 pp} (from 62.11\% to 77.37\%; \textbf{+24.57\% relative}), demonstrating practical effectiveness at scale.
\end{itemize}

The rest of the paper is organized as follows. \S\ref{sec:related-work} reviews related work. \S\ref{sec:preliminaries} formalizes the RAG setting and limitations. \S\ref{sec:method} presents \ours. \S\ref{sec:experiments} details the evaluation. \S\ref{sec:limitations} discusses limitations, and \S\ref{sec:conclusion} concludes.

\section{Related Work}
\label{sec:related-work}

\paragraph{RAG foundations.}
RAG is a prevailing approach for knowledge-intensive NLP \citep{lewis2020retrieval,borgeaud2022improving}. In the canonical pipeline, a dense retriever (e.g., \citealp{dpr}) embeds queries and documents in a shared space to fetch top-$k$ contexts, and an LLM conditions on these contexts to produce grounded answers \citep{izacard2021leveraging,atlas}. Decoupling parametric memory from non-parametric retrieval grants access to up-to-date, domain-specific information without changing model weights.

\paragraph{Improving retrieval for generation.}
One thread aligns retrieval with downstream generation via generator-aware objectives and plug-in correction \citep{shi2023replug,lin2024radit,ye2023enhancing}. A second interleaves retrieval and generation to support compositional reasoning \citep{trivedi2023interleaving,shao2023enhancing,jeong2024adaptive}. A third filters/selects context pre-decoding to improve faithfulness and efficiency \citep{wang2023learning,xu2024recomp,robustlm}. Most studies assume fixed training distributions and offline evaluation, offering limited support for non-stationary human behavior and corpus drift.

\paragraph{Instruction tuning and retrieval-augmented alignment.}
Instruction tuning equips LLMs to operate over retrieved inputs \citep{wei2022chain,wang2022self,gpt4,claude2}. Retrieval-augmented instruction/align\-ment further boosts QA and reasoning \citep{liu2024chatqa,asai2024reliable,lin2024radit,luo2023sail,wang2024rear}. However, integrating retrieval into end-to-end training requires surrogate losses and frequent re-indexing \citep{guu2020retrieval,shi2023replug,sachan2021endtoend,atlas,dong2024understand}, complicating deployment in changing environments.

\paragraph{Ranking and reranking.}
Neural ranking improves the quality of retrieved context \citep{mitra2018introduction,chen2020open}. Dual-stage and listwise architectures enhance reorderability under tight context budgets \citep{glass2022re2g,drozdov-etal-2023-parade,lin2024radit}. While common rankers rely on moderate-sized encoders (BERT/T5) that can struggle with complex semantics \citep{ram2023context}, large LLMs can act as strong rerankers with minimal prompting \citep{qin2023large,sun-etal-2023-chatgpt,khalifa-etal-2023-rerank}. Bringing this capacity to \emph{online} RAG under latency, stability, and continual-adaptation constraints remains under-explored.

\paragraph{Adaptation and feedback in RAG.}
Optimizing on static datasets implicitly assumes fixed intent and corpus. In contrast, real deployments are non-stationary, with evolving behavior and feedback. Recent works incorporate implicit/explicit feedback or self-reflection for adaptation—e.g., Self-RAG \citep{asai2024selfrag}, ReFeed \citep{refeed}, Pistis-RAG \citep{bai2024pistis}. These demonstrate the promise of feedback-driven updates but often target narrow settings or partial components, and stop short of a general-purpose interface that couples human feedback with \emph{end-to-end} retrieval control and online serving.

\paragraph{Positioning of \ours.} \ours differs along two axes. First, it \emph{structures} heterogeneous human signals—\emph{document-, list-, and response-level}—into a unified learning pipeline aligning both pointwise and listwise policies. Second, it \emph{operationalizes} this alignment for production via online fusion and distillation to a lightweight scorer, meeting latency budgets while enabling continual updates. In effect, \ours operationalizes a learnable solution for Context Engineering: it turns session-level interaction traces (retrieved contexts, their ordering, and associated feedback) into actionable supervision for retrieval control, allowing retrieval and generation modules to co-adapt during deployment and sustaining performance in open-ended, human-facing systems.

\section{Preliminaries}
\label{sec:preliminaries}

This section formalizes the RAG setting used throughout the paper and highlights structural limitations that motivate our online alignment framework.

\subsection{Problem Setup}
\label{subsec:problem-setup}

Let $\mathcal{C}=\{d_1,\dots,d_N\}$ be an external corpus. Given a human query $q\in\mathbb{Q}$, a dense retriever $R$ with encoders $(E_q,E_d)$ computes
\[
\operatorname{Rel}(q,d_i)=\langle E_q(q),\,E_d(d_i)\rangle,
\]
and returns the top-$k$ candidates
\[
D_{\mathrm{ret}}=\operatorname{Top}\!-\!k\big\{\operatorname{Rel}(q,d_i): d_i\in\mathcal{C}\big\}.
\]
A reranker $\operatorname{Rerank}_m$ (pointwise or listwise) reorders $D_{\mathrm{ret}}$ and selects the top-$m$ contexts
\[
D=\operatorname{Rerank}_m\!\left(q, D_{\mathrm{ret}}\right)=(d^{(1)},\dots,d^{(m)}),\qquad m\le k.
\]
A generator $G$ then produces a grounded response
\[
a = G(q, D).
\]
In practice, $R$ and $G$ are typically trained and deployed modularly for scalability \citep{sachan2021endtoend,atlas}.

\paragraph{Session interaction trace (retrieval \& feedback).}
In deployed systems, interactions unfold over sessions $s=(q_1,a_1,\dots,q_T,a_T)$ accompanied by human signals $\phi$ (explicit or implicit). We define the \emph{session interaction trace}
\[
\mathcal{T}_s \;=\; \big\{(q_t,\, D^{\mathrm{ret}}_t,\, D_t,\, a_t,\, \phi_t)\big\}_{t=1}^{T},
\]
as the evolving record of retrieved pools $D^{\mathrm{ret}}_t$, reranked lists $D_t$, model responses $a_t$, and human feedback $\phi_t$. We reserve \emph{working memory} to mean the token-bounded \emph{prompt} visible to the LLM at each turn. Thus, $\mathcal{T}_s$ is a \emph{log/state} over time, whereas working memory is the \emph{instantaneous} in-context input.

\paragraph{Policies and objective.}
We denote a pointwise scorer $f_{\theta}^{\mathrm{pw}}(q,d)$ and a listwise policy $\pi_{\theta}^{\mathrm{lw}}(\cdot\mid q,\mathcal{T}_s)$ that induces a permutation over $D_{\mathrm{ret}}$ and returns its top-$m$ prefix as $D$. The online objective is to adapt $\theta$ from signals in $\mathcal{T}_s$ so that the induced working memory $D$ maximizes downstream utility (e.g., satisfaction, answer quality) subject to latency and context-budget constraints.

\subsection{Limitations of Current Approaches}
\label{subsec:limitations-prelim}

Despite strong results on static evaluations \citep{lewis2020retrieval,borgeaud2022improving,guu2020retrieval}, three structural gaps hinder real-world performance:

\paragraph{(i) Fixed retrieval under non-stationarity.}
Conventional retrievers are trained offline on frozen corpora and task-specific labels (e.g., NQ, TriviaQA). At deployment they remain static, failing to reflect evolving human intent, domain drift, or long-horizon preferences.

\begin{figure*}[t]
  \centering
  \includegraphics[width=\linewidth]{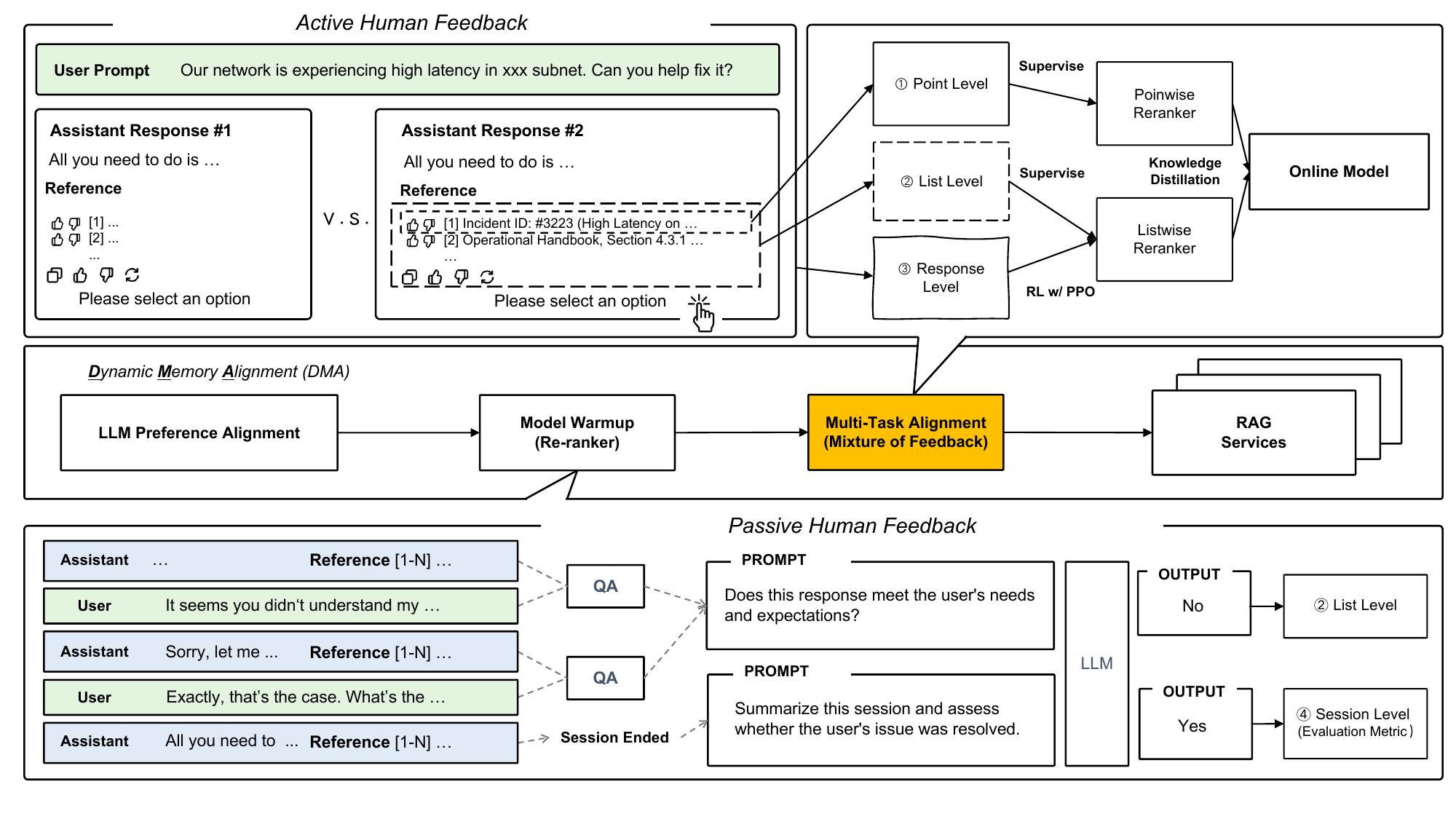}
  \caption{\ours overview. Multi-level human feedback is \emph{captured}, \emph{modeled}, and \emph{fused} to guide online retrieval/reranking. Reranker alignment and distillation for serving are detailed in Figure~\ref{fig:final-reranker-aligned}.}
  \label{fig:pipline}
\end{figure*}

\paragraph{(ii) Context bottlenecks and heuristic reranking.}
Given limited context windows, systems must aggressively prioritize evidence. Sole reliance on top-$k$ embedding similarity is brittle; rerankers/filters trained from fixed supervision often underutilize live interaction signals, reducing robustness to content drift.

\paragraph{(iii) Missing interface from feedback to retrieval control.}
Feedback at the response level (instruction tuning, preference optimization) or in browser-style agents is typically decoupled from retrieval. There is no principled mechanism to translate \emph{document-level usefulness}, \emph{list-level coverage}, and \emph{response-level satisfaction} into \emph{real-time} retrieval/reranking updates.

\medskip
As a result, online retrieval behavior remains largely frozen, inhibiting continual improvement, personalization, and adaptation to shifting distributions. These limitations motivate an online mechanism that \emph{maps multi-granularity human feedback in $\mathcal{T}_s$ to retrieval policy updates}—the focus of \ours.

\section{Dynamic Memory Alignment}
\label{sec:method}

To address the limitations identified above, we introduce \textbf{\ours}, an online learning framework that continuously adapts retrieval and reranking using real-time human feedback. \ours closes the loop between \emph{what humans signal} and \emph{how evidence is selected}. It formally \textbf{uses the \emph{session interaction trace} ($\mathcal{T}_s$)}—the evolving record of retrieved pools, reranked lists, responses, and human signals—to update retrieval control. The objective is to align the LLM's \textbf{\emph{working memory}} (the final, token-bounded context) with this feedback. Unlike static pipelines, \ours operates under deployment constraints—latency, stability, and non-stationarity—so that retrieval decisions co-evolve with human preferences while preserving robust baseline capability.

\subsection{Framework Overview}
\label{subsec:framework-overview}

As illustrated in Figure~\ref{fig:pipline}, \ours comprises three interacting modules that implement a closed feedback loop:

\textbf{(1) Feedback Taxonomy.}
\ours organizes heterogeneous interaction signals into three actionable granularities:
\emph{document-level} usefulness (pointwise signals for individual snippets),
\emph{list-level} coverage/quality (preferences over a retrieved set),
and \emph{response-level} satisfaction (pairwise or comparative judgments over answers produced from different lists).
Session-level summaries are used for evaluation and fusion weighting but not as direct supervision.
This taxonomy converts sparse, noisy, and mixed-type events into structured targets that are compatible with ranking objectives.

\textbf{(2) Preference/Reward Modeling.}
Document-level signals supervise a pointwise scorer via binary classification;
list-level signals supervise a listwise scorer via permutation-sensitive objectives;
response-level comparisons train a reward model over document lists.
The reward model provides a scalar signal that reflects downstream satisfaction for a candidate list,
enabling alignment of the listwise policy beyond heuristic list targets.

\textbf{(3) Online Adaptation \& Distillation.}
\ours aligns the listwise policy with the reward model under deployment constraints.
Concretely, the list policy is updated online (mini-batch, nearline) and then \emph{distilled}
—together with the pointwise teacher—into a lightweight gradient-boosted scorer for serving.
This preserves latency (sub-10\,ms per list) while allowing continual incorporation of fresh feedback.
Periodic refreshes maintain stability under non-stationary traffic.

Together, these modules map \textbf{session interaction traces ($\mathcal{T}_s$)} to retrieval control updates, allowing retrieval and generation to co-adapt during deployment without modifying the generator.

\subsection{Human Feedback Taxonomy}
\label{subsec:feedback-taxonomy}

Effective online adaptation requires turning heterogeneous, sparse, and noisy human interactions into learning signals that are compatible with ranking objectives. \ours structures feedback at four granularities; only the first three serve as direct supervision.

\paragraph{Document-level (pointwise) usefulness.}
humans may signal the utility of an individual snippet via explicit votes or fine-grained actions (e.g., thumbs-up/down).
We encode this as
\[
\mathcal{D}_{\mathrm{doc}}=\{(q_i,d_i,y_i)\}_{i=1}^{N_{\mathrm{doc}}},\quad y_i\in\{0,1\},
\]
optionally augmented with confidence weights $c_i\in[0,1]$ derived from agreement or heuristic reliability.
These labels supervise a pointwise scorer $f^{\mathrm{pw}}_\theta(q,d)$ via binary classification.

\paragraph{List-level coverage and quality.}
humans often react to the \emph{set} of retrieved contexts and the resulting answer.
Explicit signals include copy, regenerate or thumbs-up/down at the turn level.
We aggregate such events per query into soft list targets:
\[
\mathcal{D}_{\mathrm{list}}=\{(q_i, D^{\mathrm{ret}}_i, \tilde{\mathbf{s}}_i, \mathbf{w}_i)\}_{i=1}^{N_{\mathrm{list}}},
\]
where $D^{\mathrm{ret}}_i$ is a candidate pool, $\tilde{\mathbf{s}}_i$ encodes target utilities for items in the pool (derived from interactions that reference or expand an item), and $\mathbf{w}_i$ are per-item inverse-propensity weights (exposure correction; see \S\ref{subsec:reward-alignment}).
These supervise a listwise scorer $f^{\mathrm{lw}}_\theta$.

\paragraph{Response-level satisfaction (comparative).}
When two responses $(r_{1},r_{2})$ generated from different lists $(D_1,D_2)$ are compared, we obtain pairwise preferences
\[
\mathcal{D}_{\mathrm{resp}}=\{(q_j, D_{1,j}, D_{2,j}, y_j)\}_{j=1}^{N_{\mathrm{resp}}},\qquad y_j\in\{0,1\},
\]
indicating whether the answer from $D_{1,j}$ is preferred to that from $D_{2,j}$.
To attribute preferences to lists rather than decoding variance, we fix the generator, prompt, temperature, and decoding seed across the pair (details in \S\ref{subsec:reward-alignment}).

\paragraph{Session-level summaries (evaluation only).}
At the session granularity $s$, we compute aggregate satisfaction and dwell metrics to evaluate \ours variants (details in \S\ref{sec:experiments}).

This taxonomy (i) separates \emph{where} the signal arises (snippet, list, response, session),
(ii) specifies how to encode each signal into trainable targets (pointwise labels, listwise soft scores, pairwise preferences), and
(iii) anticipates deployment noise via confidence weighting and exposure correction.
It thus provides a consistent interface from raw interactions to learning objectives used by \ours.

\subsection{Reward Construction and Memory Alignment}
\label{subsec:reward-alignment}

\ours converts multi-granularity feedback into objectives that train (i) a pointwise scorer $f^{\mathrm{pw}}_\theta(q,d)$, (ii) a listwise policy $\pi^{\mathrm{lw}}_\theta(\cdot\mid q,\mathcal{T}_s)$, and (iii) a reward model $\mathrm{RM}(D)$ over document lists. The learning signals are designed to mitigate selection bias and decoding confounds while remaining compatible with online serving.

\paragraph{Document-level: pointwise supervision.}
Given $\mathcal{D}_{\mathrm{doc}}=\{(q_i,d_i,y_i,c_i)\}$ with confidence weights $c_i\in[0,1]$, we minimize a weighted binary cross-entropy:
\[
\mathcal{L}_{\mathrm{doc}}
= \frac{1}{Z}\sum_i c_i\Big(-y_i\log\sigma(f^{\mathrm{pw}}_\theta(q_i,d_i)) - (1-y_i)\log\!\big(1-\sigma(f^{\mathrm{pw}}_\theta(q_i,d_i))\big)\Big),
\]
where $Z=\sum_i c_i$ and $\sigma(\cdot)$ is the sigmoid. The pointwise score is later exposed as a feature to the listwise policy and the distilled scorer.

\paragraph{List-level: Pre-training from List-level Feedback.}
For $\mathcal{D}_{\mathrm{list}}=\{(q_i, D^{\mathrm{ret}}_i, S_i)\}$, where $S_i$ is a scalar score representing the quality of the \emph{entire list} $D^{\mathrm{ret}}_i$, we construct a "soft" target distribution $P_{\text{true}}$. This target $P_{\text{true}}(j)$ assumes that the list-level score $S_i$ should be attributed to items with positional decay $P_{\text{true}}(j) \propto \exp(S_i \cdot \delta_j)$, where $\delta_j$ is a positional decay factor. We then define the model distribution
$P_{\text{pred}}(j)\propto \exp(f^{\mathrm{lw}}_\theta(q_i,d_{ij}))$
and minimize the cross-entropy (ListNet) loss:
\[
\mathcal{L}_{\mathrm{list}} \;=\;
-\frac{1}{N_{\mathrm{list}}}\sum_i \sum_{j=1}^{k}
P_{\text{true}}(j)\,\log P_{\text{pred}}(j).
\]
This objective pre-trains the scorer $f^{\mathrm{lw}}_\theta$ to rank documents based on heuristics derived from list-level feedback, serving as a robust initialization for the subsequent PPO alignment phase.

\paragraph{Response-level: pairwise reward modeling over lists.}
From $\mathcal{D}_{\mathrm{resp}}=\{(q_j,D_{1,j},D_{2,j},y_j)\}$, where $y_j=1$ indicates the response generated from $D_{1,j}$ is preferred to that from $D_{2,j}$, we train a Bradley–Terry style reward model $\mathrm{RM}(D)\in\mathbb{R}$:
\[
\mathcal{L}_{\mathrm{rm}}
= -\frac{1}{N_{\mathrm{resp}}}\sum_j
\Big(y_j\log\sigma(\Delta_j) + (1-y_j)\log\sigma(-\Delta_j)\Big),\quad
\Delta_j=\mathrm{RM}(D_{1,j})-\mathrm{RM}(D_{2,j}).
\]
\textit{Confounder control.} To attribute preferences to the \emph{lists} rather than decoding variance, we fix generator, prompt, temperature, and decoding seed within each pair, and apply cross-generation for a stratified subset (swap list orders) to reduce order effects.

\paragraph{Aligning a stochastic list policy with rewards.}
We parameterize the listwise policy as a Plackett–Luce (PL) distribution over permutations of the retrieved set $D^{\mathrm{ret}}$, using item scores $g_\theta(q,d)$ (initialized from the ListNet pre-trained scorer $f^{\mathrm{lw}}_\theta$):
\[
\pi^{\mathrm{lw}}_\theta(\pi \mid q)
=\prod_{t=1}^{k}\frac{\exp(g_\theta(q,d_{\pi(t)}))}{
\sum_{u=t}^{k}\exp(g_\theta(q,d_{\pi(u)}))}.
\]
We sample permutations via the Gumbel–Top-$k$ trick, obtaining a ranked list $\pi$ whose top-$m$ prefix $D=(d_{\pi(1)},\ldots,d_{\pi(m)})$ is scored by the reward model $r=\mathrm{RM}(D)$.

We optimize $\theta$ with PPO \citep{schulman2017proximal} over a single-step episode (one list per query):
\[
\mathcal{L}_{\mathrm{ppo}}
=-\mathbb{E}\Big[\min\big(\rho_t \hat{A}_t,\,
\mathrm{clip}(\rho_t,1-\epsilon,1+\epsilon)\hat{A}_t\big)\Big]
+\beta\,\mathrm{KL}(\pi_\theta\|\pi_{\theta_{\mathrm{old}}}),
\]
where $\rho_t = \frac{\pi_\theta(a_t \mid s_t)}{\pi_{\theta_{\mathrm{old}}}(a_t \mid s_t)}$,
and the advantage $\hat{A}_t$ is derived from the list-level reward $r$ using generalized advantage estimation.

\paragraph{Nearline updates.}
To balance responsiveness and stability, \ours accumulates feedback in streaming fashion and triggers nearline updates when sufficient samples arrive (details in \S\ref{sec:experiments}). The updated list policy serves as a teacher for the online distillation stage, while the pointwise scorer provides complementary fine-grained evidence.

\begin{figure}[t]
\centering
\scriptsize
\begin{tikzpicture}[
  node distance=0.8cm and 1.6cm,
  every node/.style={font=\scriptsize, align=center, draw, rounded corners=2pt, fill=gray!5},
  box/.style={rectangle, minimum width=2.2cm, minimum height=1.0cm, text width=2.2cm, align=center},
  arrow/.style={->, thick},
  dashedarrow/.style={->, thick, dashed},
  pathlabel/.style={font=\scriptsize, draw=none, fill=none, auto}
]

\node[box, fill=white] (respfb) {Response-Level\\Feedback\\$(q, D_1, D_2, y)$};
\node[box, fill=white, above=of respfb] (listfb) {List-Level\\Feedback\\$(q, D, S)$};
\node[box, fill=white, below=of respfb] (docfb) {Document-Level\\Feedback\\$(q, d, y)$};

\node[box, right=of respfb] (reward) {Reward Model\\(pairwise; $\mathcal{L}_{\mathrm{rm}}$)};
\node[box, right=of listfb] (listtrain) {Listwise Pre-training\\(ListNet)};
\node[box, right=of docfb] (doctrain) {Pointwise Training\\(BCE; $\mathcal{L}_{\mathrm{BCE}}$)};

\node[box, right=of listtrain] (actor) {Listwise Policy Init.\\(from ListNet)};
\node[box, right=of doctrain] (pointwise) {Pointwise Scorer\\$f^{\mathrm{pw}}_\theta$};

\node[box, right=of reward] (ppo) {Policy Alignment\\PPO on PL policy};

\node[box, right=of ppo] (aligned_actor) {Aligned List Policy\\(PL; $g_\theta$)};

\node[box, below=of aligned_actor] (online) {GBDT Student\\Online Ranking};

\draw[arrow] (listfb) -- node[pathlabel] {list targets} (listtrain);
\draw[arrow] (listtrain) -- (actor);
\draw[arrow] (actor) -- node[pathlabel] {init} (ppo);

\draw[arrow] (respfb) -- node[pathlabel] {pairs} (reward);
\draw[dashedarrow] (reward) -- node[pathlabel] {list reward} (ppo);

\draw[arrow] (docfb) -- node[pathlabel] {pointwise labels} (doctrain);
\draw[arrow] (doctrain) -- (pointwise);

\draw[arrow] (ppo) -- (aligned_actor);
\draw[arrow] (aligned_actor) -- node[pathlabel] {teacher logits} (online);
\draw[arrow] (pointwise) -- node[pathlabel, swap] {teacher logits} (online);

\end{tikzpicture}
\vspace{-1ex}
\caption{\textbf{Training-to-serving pathway in \ours}
Document-, list-, and response-level feedback supervise \emph{retrieval-side} teachers. The listwise policy is aligned with the reward via PPO under a Plackett–Luce policy.
All teacher logits are then \emph{distilled into a compact GBDT scorer} for sub-10\,ms online reranking,
\emph{independent of the LLM generator}.
}
\label{fig:final-reranker-aligned}
\end{figure}
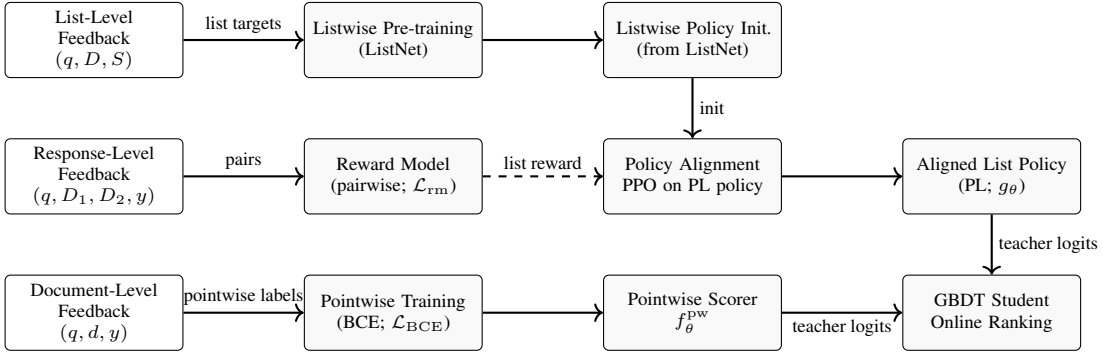

\subsection{Fusion and Distillation for Online Serving}
\label{subsubsec:fusion}

While the upstream teachers (pointwise scorer, listwise policy, and reward model) capture complementary aspects of retrieval quality, deployment requires a single low-latency ranker. \ours\ therefore \emph{distills retrieval-side teacher signals} into a lightweight scorer for online serving, entirely independent of the LLM or generation module.

\paragraph{Student model.}
The student is a gradient-boosted decision tree (GBDT) trained to predict an item-level relevance score $\hat{y}(q,d;\psi)$ for each retrieved document, and to re-rank candidates per query at serving time. It operates purely on retrieval features without any dependency on LLM representations.

\paragraph{Targets and loss.}
The distillation target is a soft relevance score $y^{\star}$ synthesized from the teacher logits:
\[
y^{\star} = \alpha\,\sigma(f^{\mathrm{pw}}_\theta)
          + (1-\alpha)\,\sigma(g_\theta),
\]
where $\alpha\!\in\![0,1]$ is a fusion weight tuned on held-out online logs.
These fused signals represent calibrated retrieval probabilities rather than response-level rewards.
The student minimizes a Huber loss to enhance robustness against outliers, trained in \emph{per-list batches} so that normalization and ordering remain consistent between training and inference.

\paragraph{Latency \& stability.}
At inference time, the GBDT independently scores all retrieved candidates and re-ranks them within each query list.
The median end-to-end scoring latency per list is under 10 ms.
Feature gating and schema versioning ensure backward compatibility when teacher models are refreshed, while shadow evaluations and incremental rollouts guard against regressions under distributional drift.

\paragraph{Summary.}
This fusion–distillation stage compresses heterogeneous retrieval-level supervision into a single, production-grade scorer that (i) satisfies strict latency constraints, (ii) preserves the listwise teacher’s relative ordering, and (iii) supports frequent nearline updates without retraining or modifying the LLM.

\section{Experiment}
\label{sec:experiments}

We evaluate \ours in two complementary settings: (1) \emph{real-world online interactions} with production humans, to validate online learning under deployment constraints; and (2) \emph{public open-domain QA benchmarks}, to assess generalization under standardized offline protocols.

\subsection{Experimental Setup}
\label{subsec:exp-setup}

\paragraph{Online evaluation (randomized controlled trial).}
We conduct a multi-month randomized controlled trial (RCT) on a GenAI assistant operated by a major telecommunications/cloud provider. Traffic is randomly split at the \emph{session} level to avoid cross-condition contamination. The control arm serves a strong, static reranker (BGE-Reranker); the treatment arm serves \ours with nearline updates. To support multilingual traffic, retrieval uses BGE-m3, and responses are generated by a fixed decoder Qwen2-72B \citep{yang2024qwen2}.

To characterize domain diversity, session queries are categorized into seven application areas: Technical support (37\%), Performance/monitoring (21\%), API/developer support (16\%), Security/compliance (10\%), Service/resource management (9\%), Migration/deployment (4\%), and Product features/updates (3\%). This distribution reflects an industrial, technically specialized environment.

\paragraph{Primary metric and label acquisition.}
Our primary outcome is \emph{session-level satisfaction}. Because explicit ratings are sparse, we infer this metric holistically for each session $s_i$ using Qwen2-72B \citep{yang2024qwen2} as an annotator with a few-shot calibrated prompt (Table~\ref{tab:prompt_template}). This evaluator LLM (distinct from any training-time models) is fed the entire session trace $(q_1, a_1, \dots, q_{n_i}, a_{n_i})$ and outputs a single categorical label $\mathbf{S}(s_i) \in \{\texttt{satisfied}, \texttt{neutral}, \texttt{dissatisfied}\}$ and a confidence score. We retain labels above a confidence threshold. Our final reported metric is the percentage of sessions where $\mathbf{S}(s_i) \neq \texttt{dissatisfied}$. Agreement with human annotations on a held-out set is high (Cohen’s $\kappa=0.962$). The prompt template (Table \ref{tab:prompt_template}) fixes role, I/O format, and exemplars to reduce drift.

\begin{table}[t]
\centering
\caption{Session-level satisfaction annotator prompt.}
\label{tab:prompt_template}
\small
\begin{tabular}{@{}p{2.8cm}@{\hskip 0.6em}p{10.2cm}@{}}
\toprule
\textbf{Intent} & \textbf{Prompt} \\
\midrule
\textbf{Role} & You are an AI assistant for evaluating human satisfaction at the \emph{session} level. \\
\textbf{Task} & Assess overall satisfaction based on the entire conversation history (queries and system responses). \\
\textbf{Input} & A session $s_i$ with $n_i$ turns: $\{(q_{i,j}, \mathbf{a}_{i,j})\}_{j=1}^{n_i}$, where $\mathbf{a}_{i,j}$ is the system's answer. \\
\textbf{Few-shot} & $\{\texttt{few\_shot\_examples}\}$ covering satisfied/neutral/dissatisfied cases. \\
\midrule
\textbf{Output} &
\textbf{Session Satisfaction}: \texttt{satisfied} / \texttt{neutral} / \texttt{dissatisfied} \\
& \textbf{Confidence}: a scalar in $[0,1]$ \\
& \textbf{Improvements}: brief suggestions to improve the experience \\
\bottomrule
\end{tabular}
\end{table}

\paragraph{Offline evaluation (public benchmarks).}
For generalization under static conditions, we evaluate on Natural Questions (NQ: 79.2k/8.7k/3.6k) \citep{nq}, TriviaQA (78.8k/8.8k/11.3k) \citep{triviaqa}, HotpotQA (88.9k/5.6k/5.6k) \citep{yang2018hotpotqa}, and WebQSP (2.8k/250/1.6k) \citep{webq}. We report Hit@1 and F1 following prior work, and \emph{fix} the decoder to LLaMA2-7B \citep{touvron2023llama2} with standardized decoding to isolate retrieval alignment effects.

\paragraph{Operational details.}
\ours’s nearline update triggers when $\sim$500 new confidence-filtered feedback instances are accumulated by a streaming pipeline (empirically balancing stability and freshness). Each cycle (i) refreshes pointwise/listwise teachers, (ii) aligns the list policy with PPO against the current reward model, and (iii) distills teachers into a 10K-tree GBDT student for serving. The end-to-end latency per update averages $\sim$10 minutes on 8 A800 GPUs; generators are served via vLLM \citep{kwon2023efficient}. Shadow evaluation precedes rollouts.

\subsection{Main Results}
\label{subsec:main-results}

\paragraph{Online RCT.}
Table~\ref{tab:api_main} reports session-level satisfaction in the randomized controlled trial.\ours improves satisfaction by \textbf{+15.26 percentage points (pp)} over a strong static reranker (BGE-Reranker) (from 62.11\% to 77.37\%; a \textbf{+24.57\% relative} improvement).We estimate uncertainty with cluster-robust standard errors at the (hashed) human level; the improvement is statistically significant. Ablations (rows B) disable one feedback granularity at a time and show that \emph{list-level} supervision contributes the largest share, followed by \emph{response-level} and then \emph{document-level}, consistent with our design hypothesis.Our GBDT distillation strategy (77.34\%) outperforms a naive cascading fusion (72.79\%) by \textbf{+4.55 percentage points} (rows C).Shifting from nearline online learning to weekly batch refreshes reduces satisfaction by 1.33 points (rows D), indicating that freshness matters even when base supervision is unchanged.

\begin{table*}[t]
  \centering
  \small
  \renewcommand{\arraystretch}{1.12}
  \setlength{\tabcolsep}{6pt}
  \caption{\textbf{Online RCT results (session-level satisfaction).}
  (A) Overall: \ours vs.\ strong static reranker.
  (B) Ablations remove one feedback granularity at a time.
  (C) Fusion: distillation vs.\ cascading.
  (D) Learning cadence: nearline vs.\ weekly.
  All CIs are cluster-robust at hashed-human level; percentages are absolute points.}
  \label{tab:api_main}

  \begin{tabular}{lccc}
    \toprule
    \textbf{Configuration} & \textbf{Satisfaction (\%)} & \textbf{95\% CI} & \textbf{$\Delta$ vs.\ ref.} \\
    \midrule

    \multicolumn{4}{l}{\textbf{(A) Overall}} \\
    \rowcolor{gray!10}
    \quad Static BGE-Reranker (control) & 62.11 & [61.64, 62.58] & Reference \\
    \quad \ours\ (full) & \textbf{77.37} & [76.99, 77.75] & \textcolor{green}{+15.26} \\
    \midrule

    \multicolumn{4}{l}{\textbf{(B) Feedback ablation (reference = full \ours)}} \\
    \rowcolor{gray!10}
    \quad \ours\ (full) & 77.37 & [76.99, 77.75] & Reference \\
    \quad w/o List-level feedback & 65.32 & [64.81, 65.83] & \textcolor{red}{--12.05} \\
    \quad w/o Response-level feedback & 68.70 & [68.21, 69.19] & \textcolor{red}{--8.67} \\
    \quad w/o Document-level feedback & 73.29 & [72.84, 73.74] & \textcolor{red}{--4.08} \\
    \midrule

    \multicolumn{4}{l}{\textbf{(C) Fusion strategy (reference = cascading)}} \\
    \rowcolor{gray!10}
    \quad Cascading fusion & 72.79 & [72.31, 73.27] & Reference \\
    \quad Distillation (GBDT student) & \textbf{77.34} & [76.95, 77.73] & \textcolor{green}{+4.55} \\
    \midrule

    \multicolumn{4}{l}{\textbf{(D) Learning cadence (reference = weekly batch)}} \\
    \rowcolor{gray!10}
    \quad Weekly batch refresh & 76.21 & [75.79, 76.63] & Reference \\
    \quad Nearline online learning & \textbf{77.54} & [77.13, 77.95] & \textcolor{green}{+1.33} \\
    \bottomrule
  \end{tabular}

  \vspace{-0.5ex}
\end{table*}

\paragraph{Offline benchmarks.}
We evaluate generalization on four public QA datasets using a fixed LLaMA2-7B reader to isolate retrieval alignment. As shown in Table~\ref{tab:offline_main}, \ours\ achieves the best Hit@1/F1 on conversational datasets (TriviaQA, HotpotQA), and remains competitive on schema-bound tasks (NQ, WebQSP). This confirms that \ours’s online alignment does not erode baseline capability and transfers to static evaluations.

\begin{table*}[t]
  \centering
  \small
  \renewcommand{\arraystretch}{1.13}
  \setlength{\tabcolsep}{5.5pt}
  \caption{\textbf{Public QA benchmarks (fixed LLaMA2-7B reader).} Left: conversational QA; Right: schema-bound QA. All numbers are \%.}
  \label{tab:offline_main}

  \begin{tabular}{lcccccccc}
    \toprule
    & \multicolumn{4}{c}{\textbf{Conversational QA}} & \multicolumn{4}{c}{\textbf{Schema-bound QA}} \\
    \cmidrule(lr){2-5} \cmidrule(lr){6-9}
    \textbf{Method} & \multicolumn{2}{c}{TriviaQA} & \multicolumn{2}{c}{HotpotQA} & \multicolumn{2}{c}{NQ} & \multicolumn{2}{c}{WebQSP} \\
    & Hit@1 & F1 & Hit@1 & F1 & Hit@1 & F1 & Hit@1 & F1 \\
    \midrule

    KnowPAT~\citep{zhang2023knowledgeable} & 63.20 & 65.20 & 29.00 & 37.40 & 51.42 & 54.82 & 68.73 & 65.31 \\
    RRHF~\citep{yuan2023rrhf}              & 62.50 & 60.20 & 28.16 & 35.40 & 50.11 & 52.01 & 66.90 & 63.10 \\
    RAFT~\citep{zhang2024raft}             & 60.10 & 57.40 & 30.20 & 35.80 & 50.24 & 53.86 & -- & -- \\
    FILCO~\citep{wang2023learning}         & 67.30 & 67.80 & 32.70 & 40.80 & \textbf{52.71} & \textbf{55.32} & \textbf{69.96} & \textbf{68.34} \\
    \rowcolor{gray!10}
    \ours\ (ours)                           & \textbf{68.81} & \textbf{68.90} & \textbf{33.92} & \textbf{41.88} & 51.11 & 54.92 & 67.26 & 65.03 \\
    \bottomrule
  \end{tabular}
\end{table*}

\paragraph{Takeaways.}
(1) Online, \ours delivers large, statistically significant gains in human satisfaction with tight latency budgets.(2) Ablation indicates a hierarchy of signal utility (list $\;>$ response $\;>$ document), matching the intuition that coarse list coverage and downstream satisfaction are most predictive for ranking.(3) Offline, \ours generalizes well and excels where conversational, multi-evidence grounding is required, while static entity-centric tasks show smaller gains—consistent with our problem framing.

\subsection{Ablation Studies}
\label{subsec:ablation-studies}

We dissect the contribution of each feedback granularity and serving design under online conditions. Unless noted, analyses use the same RCT infrastructure and report cluster-robust confidence intervals at the hashed-human level.

\paragraph{Granularity ablations.}
Starting from full \ours, we disable one supervision channel at a time:

(i) \emph{w/o list-level}: remove ListNet pre-training (list policy randomly initialized); keep pointwise and response-level supervision.

(ii) \emph{w/o response-level}: drop the reward model and PPO alignment; keep pointwise and listwise.

(iii) \emph{w/o document-level}: remove the pointwise teacher from the student.

Table~\ref{tab:api_main} (rows B) shows the largest drop when list-level signals are removed, followed by response-level and then document-level.

\paragraph{Fusion strategy.}
We compare the proposed distillation (GBDT student over teacher logits and list context) against a \emph{cascading fusion} baseline that linearly interpolates teacher scores and re-ranks.
Distillation achieves higher satisfaction (+4.55 points; Table~\ref{tab:api_main}C), lower variance across traffic slices, and better tail-latency adherence due to constant-time scoring.

\paragraph{Learning cadence.}
Switching from nearline updates to weekly batch refreshes reduces satisfaction (Table~\ref{tab:api_main}D), with larger drops during distribution shifts (e.g., product release weeks).
This indicates freshness is an independent driver of quality, even with identical supervision sources.

\section{Limitations}
\label{sec:limitations}

\ours\ focuses on aligning the retrieval layer rather than the LLM generator, which limits its ability to capture feedback related to answer style or reasoning quality. 
The reward model is list-level and may miss fine-grained factual signals. 
DMA also assumes a stable retriever backbone; major retriever updates can cause embedding drift that weakens alignment transfer. 
Finally, while nearline updates are efficient, teacher synchronization and feedback accumulation still incur nontrivial compute overhead.
Future work will explore cross-layer alignment and lighter online adaptation.

\section{Conclusion} \label{sec:conclusion}

Static RAG’s core limitation is its inability to adapt evidence selection to a human's evolving intent. We introduce \textbf{\ours}, a framework that directly addresses this by treating heterogeneous human signals as first-class supervision for controlling an LLM’s working memory. \ours contributes three components: a structured feedback taxonomy, a learning pipeline that maps signals to retrieval control, and a scalable serving recipe for real-time application.

We validate \ours using a dual-track methodology common to seminal alignment research. First, online ablations establish the causal impact of each feedback channel within an interactive regime. Second, offline few-shot evaluations confirm that the learned, human-aligned behaviors generalize robustly.

Beyond immediate performance gains, \ours's central contribution is to recast "alignment for RAG" as the \emph{control of working memory}, effectively formalizing a learnable, feedback-driven approach to Context Engineering. It provides a portable, production-ready interface to translate human feedback directly into a retrieval policy. We position \ours as a foundational step for this emerging field, enabling future work on bias-aware objectives, schema-aware retrieval, and interpretable attribution. This work moves the field beyond static competence and toward a future of continual, human-aligned intelligence.

\bibliography{ref}  
\bibliographystyle{neurips_2024}

\end{document}